# Using a Variational Autoencoder to Learn Valid Search Spaces of Safely Monitored Autonomous Robots for Last-Mile Delivery

Peter J. Bentley
Department of Computer Science, UCL, Autodesk Research
London, United Kingdom
p.bentley@cs.ucl.ac.uk

Soo Ling Lim
Department of Computer Science, UCL
London, United Kingdom
s.lim@cs.ucl.ac.uk

Paolo Arcaini
National Institute of Informatics
Tokyo, Japan
arcaini@nii.ac.jp

Fuyuki Ishikawa
National Institute of Informatics
Tokyo, Japan
f-ishikawa@nii.ac.jp

## ABSTRACT

The use of autonomous robots for delivery of goods to customers is an exciting new way to provide a reliable and sustainable service. However, in the real world, autonomous robots still require human supervision for safety reasons. We tackle the real-world problem of optimizing autonomous robot timings to maximize deliveries, while ensuring that there are never too many robots running simultaneously so that they can be monitored safely. We assess the use of a recent hybrid machine-learning-optimization approach COIL (constrained optimization in learned latent space) and compare it with a baseline genetic algorithm for the purposes of exploring variations of this problem. We also investigate new methods for improving the speed and efficiency of COIL. We show that only COIL can find valid solutions where appropriate numbers of robots run simultaneously for all problem variations tested. We also show that when COIL has learned its latent representation, it can optimize 10% faster than the GA, making it a good choice for daily re-optimization of robots where delivery requests for each day are allocated to robots while maintaining safe numbers of robots running at once.

## 1 INTRODUCTION

Home delivery of groceries and other goods is rapidly becoming a major new industry worldwide, accelerated by the COVID pandemic. The use of automobiles for last-mile delivery (from store to customer) causes environmental concerns [1] and in countries such as Japan where our problem originates, there may be a lack of labor. Our industry partner Panasonic is currently trialing their solution: an automatic delivery service performed by autonomous robots. Customers order goods which are then collected from a supermarket or drug store by a robot and delivered to the customer. The robots are autonomous but have human monitors that intervene should a problem occur, such as a potential collision with a pedestrian, Figure 1. The use of robots for this purpose is in the experimentation phase, with details such as number of robots and human monitors still under consideration.

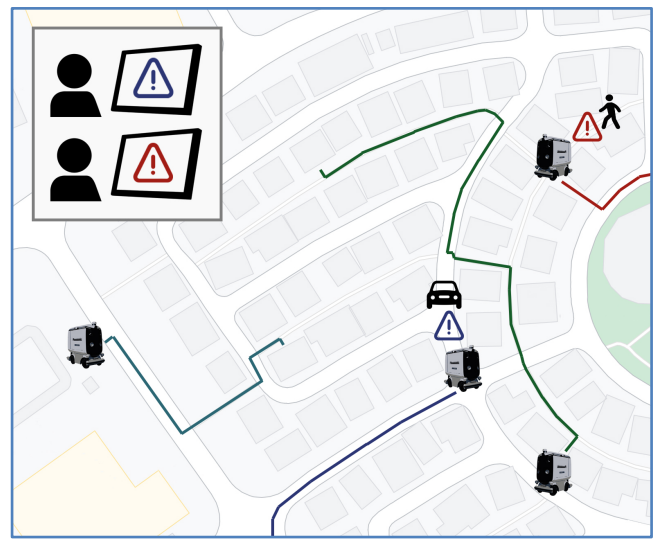

**Figure 1: Home delivery service using low speed robots with human operators performing safety monitoring. Map shows section of actual location.**

In this real-world problem there is a clear conflict between fulfilling orders and maintaining safety. More robots operating simultaneously will mean more orders could be fulfilled, yet too many robots operating at once would be dangerous as the human operators can only monitor a limited number of robots. The problem of how best to schedule different numbers of robots such that maximum orders are delivered safely is therefore difficult. It is also a problem that must be solved repeatedly for every new set of customer orders. This problem is likely to become increasingly relevant as autonomous robot deliveries become more prevalent, yet safety is rarely considered in the literature.

We address this problem by using a hybrid machine learning and evolutionary computation approach [2, 3], designed to improve constrained optimization without relying on carefully crafted operators or weighting factors. This method uses a variational autoencoder to learn the valid regions of the search space, i.e., the areas of the search space where safe numbers of robots are scheduled to work simultaneously. A genetic algorithm (GA) is then used to search in this learned latent space to find optimal timings for robots such that the maximum number of orders can be delivered safely. We compare this approach with a baseline genetic algorithm optimizing robot timings while solving the constraint, without the use of machine learning. We investigate how well both approaches enable us to examine different design variations and we assess suitability for long term use once the design variables are determined.

The contributions of this work can be summarized as follows:
- The first study to explore real-world autonomous robot routing while satisfying a safety monitoring requirement.
- Adapting the hybrid ML-EA approach (COIL) [2, 3] to a real-world problem for the first time.
- Evaluation of COIL by comparing to a baseline GA to explore problem variations, showing better solution quality and optimization speed for daily robot scheduling.
- COIL performance is enhanced on the problem through reduction of the training set and number of latent variables.

The rest of the paper is organized as follows: Section 2 provides a literature review of relevant research, Section 3 describes the method, Section 4 describes the experiments, results and discusses findings for the problem. We conclude in Section 5.

## 2 BACKGROUND

### 2.1 Autonomous Robots for Home Delivery

Autonomous delivery robots have the potential to significantly reduce energy consumption and $CO_2$ emissions in urban areas [1]. There are several examples of solutions. Mercedes-Benz Vans partnered with Starship Technologies to propose the concept of autonomous delivery robots launched from trucks [4]. The truck can replenish robots at decentralized depots to launch more until all its customers are supplied. Boysen et al. [4] developed scheduling procedures which determine the truck route, such that the weighted number of late customer deliveries is minimized.

Yu et al. [5] considered a variant of this problem and proposed an LV-SAV (large vehicle - small fully automated ground vehicles) model where multiple LVs cooperate with their associated SAVs. Safety was one of their concerns, and having slow moving SAVs is safer in urban delivery, and was considered to be safer than using drones in the HVDRP (hybrid vehicle-drone routing problem) proposed by Karak and Abdelghany [6].

Chen et al. [7] introduced another variant called a new vehicle routing problem with time windows and delivery robots (VRPTWDR). They showed that considerable operational time savings can be achieved by dispatching delivery robots to serve nearby customers while a driver is also serving a customer.

Since delivery robots share sidewalks with pedestrians, Bakach et al. investigated the option to choose paths that avoid zones with high pedestrian density [8]. They considered path flexibility given the presence of zones with varying pedestrian level of service.

The study of vehicle routing problems (VRP) continues to develop, with many variants including real-life constraints and assumptions, which makes the models more realistic and the approaches more applicable in practice [9]. Autonomous delivery vehicles introduce safety concerns whilst driving autonomously on the public road networks and performance risks while delivering parcels (e.g., risk of malfunctioning of the technology) [10]. A study of public perceptions and acceptance of autonomous vehicles found that people expressed high concerns around safety, cyber-security, legal liability, and regulation issues [11].

In this work we focus on safety and a key component of the solution created by Panasonic [12, 13]. New to this field, for our problem, human operators monitor the robots so that they can manually operate them when unexpected obstacles are detected, or they can talk with the users. Similar monitoring will be required for many other real-world examples, so methods to solve this problem are likely to be widely useful.

### 2.2 Evolving Latent Variables

Autoencoders [14] are neural networks first used for dimensionality reduction. Since their inception they have become popular for learning generative models of the data. An autoencoder comprises two parts - an encoder $p$, which maps the observations $x$ to a (lower dimensional) embedding space $z$ and a decoder $q$, which maps the embeddings back to the original observation space. When trained, the autoencoder minimizes the error in reconstruction of output compared to input. The variational autoencoder (VAE) is a probabilistic autoencoder [15, 16]. Instead of encoding an observation as a single point, VAEs encode it as a distribution over the latent space.

Recent work in evolutionary computation makes use of VAEs to learn representations and then evolve using them, searching in the latent space (LVE – latent variable evolution). Game levels [17], fingerprints to foil security systems [18], and program synthesis [19] include some of the applications. Researchers also make use of quality-diversity approaches [20] with LVE to improve optimization in high dimensional problems, for example DDE-Elites [21] which uses a combination of the direct encoding with a learned latent encoding to accelerate search.

Most recently, COIL (constrained optimization in learned latent space) [2] and SOLVE (search space optimization with latent variable evolution) [3] present an LVE approach to learn a latent representation that is biased towards solutions that satisfy a constraint or additional objective. This is very relevant to the real-world problem we investigate in our work. However, while COIL [2] and SOLVE [3] have been demonstrated on simple constraints and benchmark functions, the technique has not been tested on a real-world problem. In this work we apply the concepts in COIL for the first time to the problem of scheduling a safe number of autonomous robots for home delivery of a maximum number of orders by customers.

## 3 METHOD

We base our approach on the system under development by our partner Panasonic. Their automated delivery service will enable customers to order goods from a local store which are delivered by autonomous robots. Customers submit requests through an app. Each request is scheduled centrally with a robot allocated to serve it. Robots are monitored by human operators who can intervene if the robot encounters problems, helping to initiate the appropriate response in the robot or talk to users, Figure 1.

We focus on two stages to this problem:
1. Investigate possible design variations, with different numbers of robots, human operators, request durations.
2. On completion of the first stage, provide an efficient method for daily optimization of robot timings.

For the first stage, we investigate optimization methods capable of exploring design variations. The algorithms must successfully schedule the robots to maximize delivery of requests while ensuring there are never too many robots running simultaneously, such that the human operators can safely monitor them. For the second stage, with the design variables now determined, we investigate the fastest method of scheduling robots to meet the human operator constraint, daily, because this decision-making process reoccurs every day with new customer requests.

We formulate the problem as a constrained optimization problem. Given a randomly generated set Reqs comprising $Rq$ customer requests, each with duration $Rqdur$:

Reqs = {$Rqdur_1$, .., $Rqdur_{Rq}$}

where $Rqdur_j$ = random[60..$dr$] and given a set of robots Rbts comprising $Rb$ pairs of robot starting time Rst and running time Rrt:

Rbts = {($Rst_1$, $Rrt_1$), .., ($Rst_{Rb}$, $Rrt_{Rb}$)}

allocate each request $Rqdur_j$ in order from $j$ = 1 to $Rq$ (i.e., on a first-come, first-served basis) to the most appropriate robot using a simple scheduler created for this work, Algorithm 1. (Should this solution be commissioned for use on the real robots, this algorithm will be replaced with the industrial partner's internal scheduler.)

**Algorithm 1: First-come first-served scheduler used to allocate customer requests to robots. *totalRm* counts the total number of requests met out of a maximum of *Rq*.**

```
totalRm = 0
for i = 1 to Rb
    remaining_dur_i = Rrt_i×10
endfor
for j = 1 to Rq
    if there exists a remaining_dur_k such that
        0 <= ( remaining_dur_k - Rqdur_j ) < 10 and
        there are no other closer matches then
            remaining_dur_k  -= Rqdur_j
            totalRm ++
            break
    else find largest remaining_dur_k such that
        ( remaining_dur_k - Rqdur_j ) > 0 then
            remaining_dur_k  -= Rqdur
            totalRm ++
            break
endfor
```

The task is then to use an optimizer to find a suitable set of Rbts such that the number of requests fulfilled by the robots is maximized by efficiently fitting robot start times and running times to the set Reqs while ensuring that the number of robots running simultaneously at any point in time remains below the maximum robot threshold $RT$.

We assume that robots operate 12 hours in a day (e.g., 7am to 7pm) and they run at a fixed speed. The minimum running time for a robot is one hour. Robots only operate for a single period per day. We divide the 12-hour period into 10-minute time slots such that every start time and duration for each robot may be defined by an integer from 0 to 66, where $Rst_i$ = 0 corresponds to a robot start time of 7am and $Rrt_i$ = 0 corresponds to the minimum duration of 1 hour. The set of requests Reqs is built from $Rq$ random integers from 60 to 180 (i.e., customer requests take robots anything from 60 to 180 minutes to fulfil). Customer requests must be fulfilled within the 7am to 7pm working period of robots and goods may be delivered at any point during that period for all requests. By default $Rq$ = 120 to present a significant challenge for the robots; for experiments with shorter request durations we double $Rq$ to 240.

In this work we measure the success of solutions for the problem using the objective function to be maximised:
$$f(\bar{x}) = totalRm$$
where *totalRm* is the total number of requests from Reqs allocated by the scheduler to the current set of robots Rbts.

We use a single constraint:
$$totalRbts \leq RT$$
where *totalRbts* is the total number of robots working simultaneously in any one timeslot. As a human operator must monitor each robot in this problem, $RT$ directly correlates with the number of human operators. Figure 2 illustrates the calculation of *totalRbts*.

| Rbts | 0-10 | 10-20 | 20-30 | 30-40 | 40-50 | 50-60 | 60-70 | 70-80 | 80-90 | 90-100 | 100-110 | 110-120 |
|---|---|---|---|---|---|---|---|---|---|---|---|---|
| 1 | 0 | 0 | 0 | 1 | 1 | 1 | 1 | 0 | 0 | 0 | 0 | 0 |
| 2 | 0 | 0 | 0 | 0 | 1 | 1 | 1 | 1 | 1 | 1 | 1 | 1 |
| 3 | 1 | 1 | 1 | 1 | 1 | 0 | 0 | 0 | 0 | 0 | 0 | 0 |
| 4 | 1 | 1 | 1 | 1 | 1 | 1 | 1 | 1 | 1 | 1 | 1 | 1 |
| + | 2 | 2 | 2 | 3 | 4 | 3 | 3 | 2 | 2 | 2 | 2 | 2 |

**Figure 2: Example calculation showing *totalRbts* = 4, for four robots in a two-hour period (ignoring the minimum duration of 60 for this example). If the threshold *RT* = 2 then the constraint would not be met on 4 out of the 12 possible timeslots, marked in bold. When *totalRbts* <= *RT* for every timeslot for all robots, the constraint is met fully. In this case the constraint score being minimized is 4 - 2 = 2. With the constraint not satisfied, this is not a valid solution.**

### 3.1 Optimization with Genetic Algorithm

Our baseline approach to address this problem is to use a genetic algorithm to evolve start times and durations for a set of robots such that the constraint is met and the objective is maximized for a given set of customer requests *Reqs*.

Our search variables $\bar{x}$ comprise:
$$\bar{x} = [(x_{st}^1, x_{rt}^1), \ldots, (x_{st}^{Rb}, x_{rt}^{Rb})]$$
where we map $(x_{st}^i, x_{rt}^i)$ to ($Rst_i$, $Rrt_i$) for every *i*. Should any *i* exist such that $x_{st}^i + x_{rt}^i > 65$ then the corresponding values of $Rrt_i$ is corrected such that the robot ceases its shift at the end of the working day. Values are corrected during this mapping stage rather than modifying the evolving parameters as this has been shown to be more conducive to search [22, 23]. During evolution the fitness and constraint is calculated as described above. This is the worst-case scenario for the constraint as it assumes that every robot will run for its maximum permitted time. In reality, after requests have been scheduled some robots may run for shorter durations depending on which requests were allocated. Thus, for accuracy, our reported results for all experiments first updates robot durations according to the scheduler in Algorithm 1:
$$\forall i \in \{1..Rb\} \; Rrt_i \mathrel{-}= \lceil remaining_{dr_i}/10 \rceil$$
and then measures the actual number of scheduled robots that ran simultaneously, as shown in Figure 2.

We use a constrained optimization algorithm [24] using DEAP [25] with tournament fitness (see [2] for the algorithm) to enable the equal contribution of constraint and objective to evolution – a common approach in constraint-handling [26]. In this method, fitness of each individual is the number of times its separate criteria win a series of tournaments held between randomly selected subgroups of 3 individuals. Within each tournament, the fitness of an individual is increased if its solution is better (lower) for the number of times the constraint is not satisfied, and when it is better (higher) for the objective $f(\bar{x})$ compared to the others in the tournament. We use an integer version of Gaussian creep mutation and uniform crossover [25].

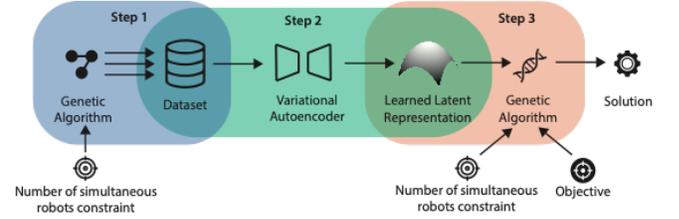

**Figure 3: Constrained optimization in learned latent space (COIL) applied to real-world autonomous robot scheduling problem.**

### 3.2 Optimization with COIL

We are tackling a challenging constrained optimization problem, so our proposed approach to be compared with the baseline is COIL [2] – a new hybrid machine learning and optimization method designed to learn valid regions of the search space and improve the ability of optimizers to find valid solutions. COIL has the advantage that, unlike other constrained optimizers, it does not rely on complex operators, weight-tuning, or specialized expertise. So far COIL has been tested only on simple test functions for constrained optimization; this is the first research to apply this approach to a real-world problem.

COIL operates using three steps (Figure 3):
1. Dataset generation from constraint
2. Learning new latent representation from dataset
3. Optimization using latent representation using constraint and objective

For the first step, we use a simple genetic algorithm (using DEAP) with a population size of 200 for running for at most 200 generations. Our search variables $\bar{x}$ are exactly as described for the baseline GA. The objective function for this data-generation GA is minimization of the number of times the constraint is not satisfied, with any solution that satisfies the constraint added to the dataset, and the GA restarting. The threshold *RT,* which determines how many robots may work simultaneously, is varied during experiments. Solutions that specify robot durations that finish beyond the 12-hour working period are corrected as described in 3.1 before being normalized and added to the dataset. To encourage the dataset-generating GA not to "cheat" and only provide solutions with minimal robot running times (the simplest way to satisfy the constraint), we also add a selective pressure towards longer robot durations by adding $100/\sum_1^{MR} x_{rt}^i$ to the fitness term. We run this GA repeatedly until *DS* vectors are generated. Each vector represents a valid set of robot start times and durations such that the constraint is satisfied. The overall objective to find robot times and durations that enable the most customer requests is not used at this point.

In the second step, we provide the dataset to a simple variational autoencoder [2] with 4 linear layers, a prior of *N(0,I)*, KLD = 1, for 200 epochs using the Adam optimizer with a learning rate of 0.001. We run the VAE 10 times and choose the learned model with the lowest error. This is our learned latent representation, with latent variables:
$$\bar{z} = [z^1, z^2, .., z^{2 \times maxlv}]$$

In the third step, we use the same GA as described in section 3.1, to ensure a completely fair comparison. For this GA, the search variables $\bar{z}$ are encoded as real-valued variables with ranges between -2.0 and 2.0, and a Gaussian creep mutation. We convert $\bar{z}$ into $\bar{x}$ by using the learned VAE model to express the values, and then perform a scalar inverse transform, unnormalization, and conversion to integer to convert the VAE output into the desired 0..66 range. We then map $x_{st}^i, x_{rt}^i$ to $Rst_i$, $Rrt_i$ respectively for every $i$ as described in section 3.1 with the same robot duration updates performed according to the scheduler and use the identical tournament fitness approach to measure fitness for the objective and constraint.

## 4  EXPERIMENTS

### 4.1  Experiments

We perform experiments to investigate the following research questions, which reflect the two stages to this problem:

*RQ1: Does COIL help us explore valid design variations more effectively than the baseline GA?*
We explore this question in several experiments.

*E1.1: Can the baseline GA and COIL help us explore valid solutions?* Before we can explore different design variations we need to check that we can solve this problem at all. Experiment 1 tackles this by directly comparing the output of the baseline GA with COIL. Following the training of the VAE within COIL, the same learned model is used in all repeated runs. We use a small population size of 20 for just 50 generations following [2, 3]. Both GAs are run 100 times and mean results are shown. A new set of requests *Reqs* is randomly generated for every run.

*E1.2: Can the GA and COIL explore valid solutions varying RT?* In this experiment we explore variants of the problem by varying *RT* to the values: 10, 15, 20. This explores solutions where we permit more robots to run simultaneously (at the cost of requiring more human monitoring.)

*E1.3: Can the GA and COIL explore valid solutions varying Rb?* In the second problem variant, we fix *RT* to 10 and vary the number of robots *Rb* to the values: 20, 25, 30. This explores solutions where we have fewer robots overall, with at most 10 running simultaneously.

*E1.4: Can the GA and COIL explore valid solutions varying dr?* In our final set of problem variants, we vary the duration of requests; instead of random from 60 to 180, we try 60 to *dr*, where $dr \in \{60, 80, .., 360\}$.

For all RQs, parameters other than those being varied remain unchanged. COIL is run 100 times for each setting, with a new set *Reqs* of random requests each run.

*RQ2: Is COIL an efficient algorithm for daily optimization?*
Once the company decides on the number of robots and operators to use, we arrive at the second stage. Here we investigate the suitability of COIL for use as a daily optimizer, investigating whether the method can be tuned with this goal in mind.

*E2.1: Can we use smaller dataset sizes for COIL without affecting its ability to generate useful latent representations?* We need a fast way of optimizing robot schedules every day for the new set of requests, while always meeting the constraint. One drawback of COIL is the need to generate a dataset first and train the VAE. Although this is a one-off, offline computation, it is still significant. This experiment addresses this by varying the dataset size *DS* to four different values: 2500, 5000, 7500, 10000. All other parameters remain unchanged. COIL is run 100 times (with a new set *Reqs* of random requests each run) and mean results are shown. We also measure the difference in data generation and training times and the error rate of the VAE.

*E2.2: Can we improve the performance of COIL by reducing the number of latent variables?* Here we investigate an idea for improving the performance of COIL. While COIL was described as a method to improve the search space while keeping the number of latent variables the same as the number of input variables, other work has shown that VAEs may be able to provide the additional benefit of reducing the search space size [27]. We investigate this question with the number of robots *Rb* = 30 (giving us 60 problem variables) by varying *maxlv* to the values: 5, 10, 15, 20, 25, 30 (i.e., 10 to 60 latent variables). All other parameters remain unchanged. COIL is run 100 times for each setting, with a new set Reqs of random requests each run.

Table 1 summarizes the parameters investigated in each experiment. All processing was performed on a MacBook Air 2020 M1 with 16Gb memory. All code was implemented in Python and is available[1].

**Table 1. Parameters for the experiments**

| Exp | DS | maxlv | RT | Rb | dr |
|---|---|---|---|---|---|
| E1.1 | 10000 | *30* | 10 | 30 | 180 |
| E1.2 | 10000 | *30* | 10,15,20 | 30 | 180 |
| E1.3 | 10000 | 30 | 10 | 20,25,30 | 180 |
| E1.4 | 10000 | *30* | 10 | 30 | 60..360 |
| E2.1 | 2500..10000 | *30* | 10 | 30 | 180 |
| E2.2 | 10000 | 5..30 | 10 | 30 | 180 |

### 4.2  Results and Analysis

In this section we describe the results for research question RQ1 (experiments E1.1, E1.2, E1.3 and E1.4) and for RQ2 (experiments E2.1 and E2.2).

**Table 2. E1.1: COIL vs GA. Better results in bold.**

|  | COIL | GA |
|---|---|---|
| avg objective (stdv) | 66.43 (6.80) | **93.6 (5.55)** |
| min objective | 51 | **76** |
| max objective | 77 | **107** |
| avg constraint (stdv) | **23.93 (8.43)** | 44.7 (3.89) |
| min constraint | **0** | 35 |
| max constraint | **30** | 52 |

---

[1] https://github.com/writingpeter/coil_gecco23

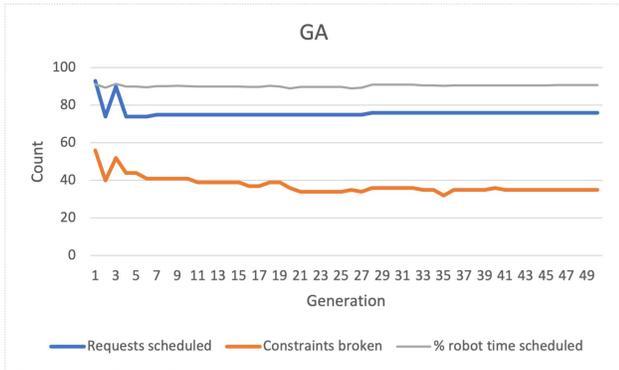

**Figure 4: Example single best run baseline GA.**

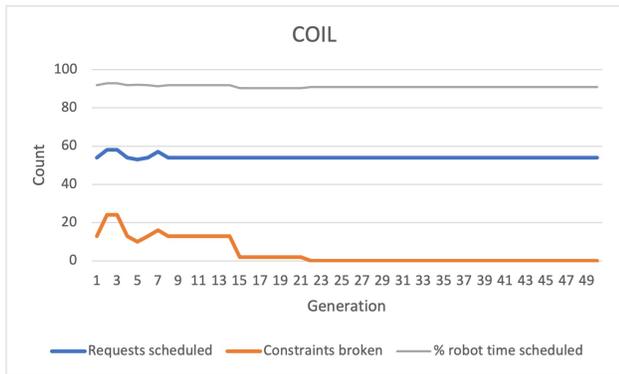

**Figure 5: Example best run COIL.**

### E1.1: Can the GA and COIL help us explore valid solutions?

In this experiment we compare the output of the baseline GA with COIL to check that both can find valid solutions for our default parameter values. Table 2 summarizes the results. The results show that only COIL is able to provide valid solutions for this variant of the problem. COIL provides superior results in terms of constraint satisfaction, with the best results being a perfect zero, and average of 23.9, compared to the baseline GA best of 35 (worse than the worst score of COIL) and average of 44.7. The scores for the objective shows that GA allocates more tasks on average than COIL, managing 107 at best compared to COIL's much lower 77. However, the scores are meaningless when the constraint is broken – the baseline GA cheats by not meeting the constraint in order to allocate more requests, and such solutions are not viable or useful.

Examining a single best run for the baseline GA and COIL, Figure 4 and Figure 5 show the difference in evolution. The GA attempts to improve the constraint while keeping request allocation constant but never achieves a single valid solution – showing a failure of this algorithm to tackle the problem effectively. In contrast COIL starts better and optimizes the constraint to zero very rapidly while keeping the number of requests allocated constant. To assess whether the GA could be coerced into focusing more on the constraint, we increased the weighting by 10 times in the tournament selection algorithm for the constraint; we also tried much larger population sizes and number of generations. The results were barely changed and the GA still failed to find any solutions that satisfied the constraint. Figure 4 and Figure 5 (grey line) also show how effectively $\bar{x}$ is being optimized, with both algorithms ensuring that 91% of the robots' time is being used. While COIL does not find perfect solutions every time for this difficult problem (unlike findings of [2, 3]) it is clear that its learned latent representation is biased towards solutions that satisfy the constraint more, with Figure 6 top left showing that the baseline GA results (orange circles) are all worse than the COIL results (blue circles).

### E1.2: Number of simultaneously running robots *RT*

When we increase the value of *RT*, we make the constraint easier as more robots are permitted to run simultaneously. The results show that in all problem variants, COIL successfully finds valid solutions to the problem. However, as the optimization problem becomes easier, the baseline GA begins to function more usefully (Figure 6 top left). When *RT* = 10, COIL performs better than the baseline GA. When *RT* = 15, COIL is able to find more solutions that satisfy the constraints, while the baseline GA still finds none. But when *RT* = 20 (the problem is no longer difficult) there is no longer any significant difference between the GA and COIL, with COIL finding slightly more valid solutions, but the GA finding a few solutions that satisfy the constraint and fulfil the objective slightly better.

### E1.3: Total number of robots *Rb*

Similar to the previous experiment, when we reduce the value of *Rb*, we make the constraint easier as fewer robots in total are running, so fewer are likely to run simultaneously. However, with *RT* fixed we do not see improvements in objective scores (keeping the number of simultaneously running robots constant, limits the number of requests that can be served). When reducing *Rb* from 30 to 25, the baseline GA comes closer to finding valid solutions. When *Rb* = 20, the GA finds several valid solutions. However, for all values tested, COIL always finds valid solutions to the problem, becoming more consistent as the problem is easier (Figure 6 top right).

### E1.4: Changing request durations *dr*

If we alter the request durations (increasing the number of requests from 120 to 240 to make up for smaller durations), we change the difficulty of the problem in a new way: shorter tasks are easier to allocate while longer tasks are more difficult. Figure 7 shows how the average number of constraints broken slightly improves (lowers) as the request duration increases for COIL, suggesting that COIL scales better to greater problem difficulty. As expected, the number of requests that can be allocated falls as the task durations increase, with the baseline GA always allocating more (because it cheats by not satisfying the constraint).

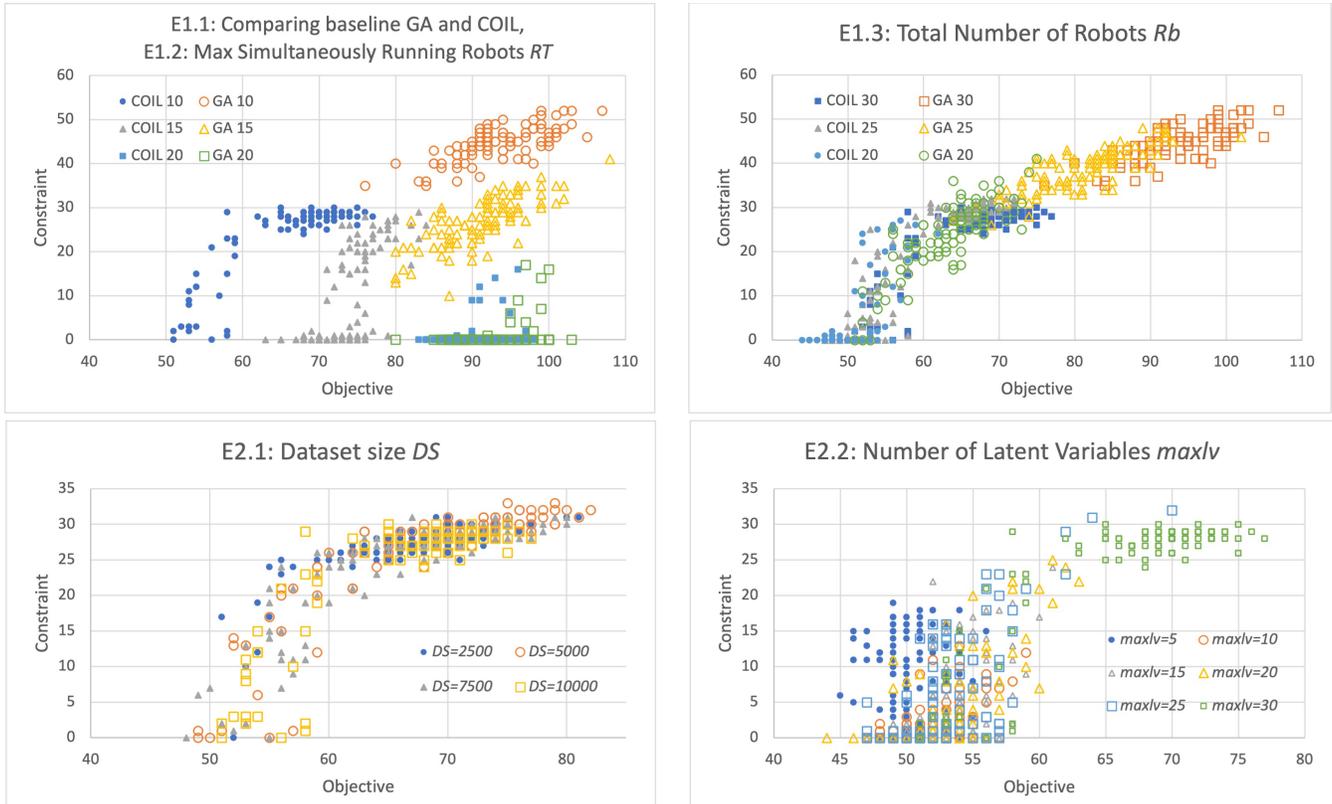

**Figure 6: Best result at final generation for 100 runs for each algorithm, plotted with objective score against constraint. Lower means constraint is better satisfied, more to the right means more tasks are allocated. Only solutions on y = 0 satisfy the constraint fully and thus are valid. Top left: E1.1 Comparing GA and COIL and E1.2 Varying *RT*. Top right: E. 1.3 Varying *Rb*. Bottom left: E2.1 Varying COIL *DS*. Bottom right: E2.2 Varying COIL *maxlv*.**

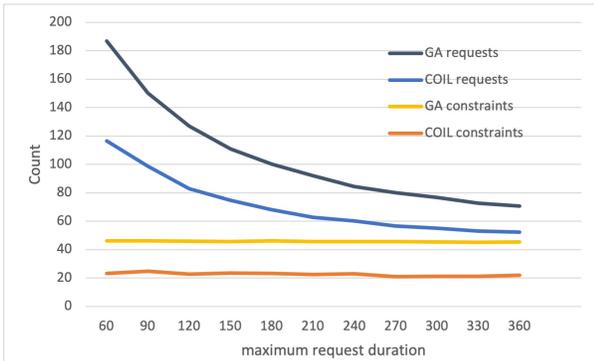

**Figure 7: E1.4: Altering maximum random request durations.**

### E2.1: Reducing dataset size *DS*

The previous experiments showed that COIL clearly outperforms the baseline GA for this real-world problem – only COIL has permitted us to explore valid solutions (the constraint fully satisfied) for all design variations tested. However, there is a difference in computation time required. When evolving the solutions for E1.1, the baseline GA took 1.89 minutes to complete 100 runs. COIL took a slightly shorter 1.81 minutes – despite the additional need to express the evolving latent variables into the problem variables using the VAE. However, COIL also required a one-off pre-computation to generate valid data and use the VAE to learn the latent representation. This computation was substantial, requiring 25.80 hours to generate 10,000 datapoints, and 8.95 minutes for the VAE to learn (running 10 times and choosing the best learned model, which had loss of 0.9114). These times reduced when the problem was easier, for example in E1.2, 10,000 valid points were generated in just 19.28 minutes when *RT* = 15, and just 8.08 minutes when *RT* = 20.

In the next experiment, we investigate the effects of the dataset size to see if COIL can still achieve acceptable results when the VAE is trained with smaller datasets. We achieved this by simply running the data generator each time with *DS* = 2500, 5000, 7500, and 10000. The datasets were checked to see if they contained any redundancy through duplication; analysis revealed that every point in each dataset is unique and not repeated.

Figure 8 shows the differences in computation time vs the VAE loss, indicating that the dataset of size 5000 has the smallest loss; detailed results from COIL for the constraint and objective for each dataset size are provided in Supplementary Materials. While the best average constraint satisfaction is achieved using the largest dataset, performance is still relatively unaffected for even the smallest size. Figure 6 (bottom left) shows that valid solutions on *y* = 0 are generated by all dataset sizes, and all also have a similar trend in the solution space.

**E2.2: Modifying number of latent variables *maxlv***

Finally we investigate whether we can improve the performance of COIL in terms of the quality its solutions. Applying COIL with different numbers of latent variables shows an important effect on the distribution of solutions, Figure 6 (bottom right) and Supplementary Materials. Reducing the number of latent variables generally improves the ability of COIL to find solutions that satisfy the constraint. Fewer latent variables mean faster training time for the VAE, and faster optimization for the GA using the smaller latent representations. But as the number of latent variables decreases to 10, the VAE loss becomes worse (Figure 9). The smaller search space also appears to impact the objective, with fewer tasks being allocated.

In contrast, increasing to 50 or more latent variables appears to provide an unwelcome bias away from valid solutions on $y = 0$. For this problem, there appears to be a "sweet spot" between *maxlv*=15 and *maxlv*=25 (30 and 50 latent variables respectively), where solutions with high number of request allocations and perfect constraint scores are generated (Figure 6 bottom right). For our problem it appears that having fewer latent variables compared to problem variables is advantageous.

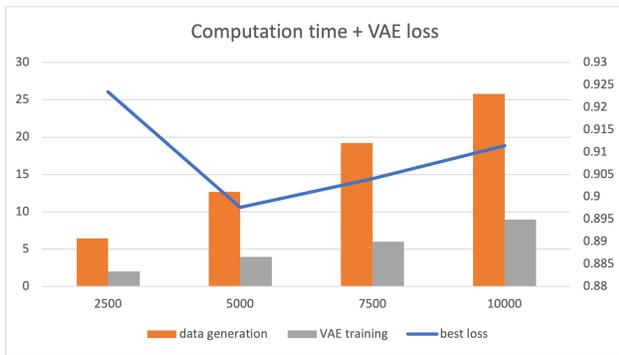

**Figure 8: Computation time and VAE loss for E2.1. Bar chart using left *y*-axis: Data generation (hours), VAE training (minutes), blue line using right *y*-axis: VAE loss.**

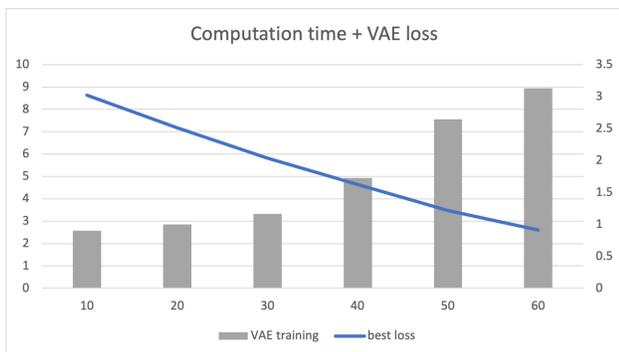

**Figure 9: Computation time and VAE loss for E2.2. Bar chart using left *y*-axis: VAE training (minutes), blue line using right *y*-axis: VAE loss.**

### 4.3 Discussion of Findings Relating to Problem

COIL has enabled a useful exploration of the design space for this real-world problem. Our findings indicated that the amount of available time spent by robots delivering requests rarely changes (usually at around 90%). This appears to be the best use of the robots' time as determined by the scheduler, and both optimizers are always able to find appropriate start times and durations for the autonomous robots to reach this efficiency. However, results from the baseline GA were often unusable as the safety constraint was not met. COIL provided usable results for all setups. The experiments also showed that when customer requests were likely to take less time, more would be delivered successfully by the robots. The safety constraint was slightly more likely to be satisfied by COIL for longer duration requests; for the baseline GA there was no difference.

Overall, COIL's exploration of the problem space indicates that when we keep the total number of robots *Rb* constant but allow more to run simultaneously (i.e., more human monitors are employed), the number of customer requests that can be satisfied will increase. In contrast, if we keep the number of simultaneously running robots *RT* constant and reduce the overall numbers of robots that are running, there is little change to the number of customer requests met. Thus, with the fixed relationship between the number of simultaneously running robots and the number of human monitors, there appears to be a direct correlation between the number of human monitors and customer satisfaction.

## 5 CONCLUSIONS

The problem of scheduling autonomous robots safely such that sufficient human monitors are available is challenging, and relatively unexplored in the literature. Our work shows that a standard genetic algorithm using a well-established method for constrained optimization (the baseline GA) was unable to find solutions that satisfied the constraint except for variations where the problem was relatively easy. In contrast a new ML-EA hybrid approach (constrained optimization in learned latent space: COIL) was able to find valid solutions for all problem variants, enabling us to perform a useful investigation of the effects of each parameter on solutions. This work has also shown for the first time that the data-generation and training stage of COIL can be improved by reducing the dataset size and number of latent variables. This means that once suitable parameters are selected, and the initial one-off training is performed to learn the latent representation, COIL is able to optimize new schedules rapidly and reliably. Indeed, with these improvements, COIL is 10% faster to run than the baseline GA. COIL is thus an ideal choice for an everyday optimizer for this problem.

Future work will examine further problem variables of interest to our industrial partner such as robot speed and will integrate with other schedulers. Other improvements to COIL will be examined, such as the use of more advanced VAEs and optimizers such as MAP-Elites for the data generation and optimization stages, to further improve speed and efficiency.

## ACKNOWLEDGMENTS
We thank Hirokazu Kawamoto, Kaoru Sawai, and Eiichi Muramoto of Panasonic Holdings Corporation for the discussion on autonomous delivery robots.

# SUPPLEMENTARY MATERIALS

**Table S1: E1.2: Modifying *RT:* number of simultaneously running robots, COIL.**

| COIL | 10 robots at once | 15 robots at once | 20 robots at once |
|---|---|---|---|
| avg objective (stdv) | 66.43 (6.80) | 73.94 (3.99) | 89.86 (3.14) |
| min objective | 51 | 63 | **83** |
| max objective | 77 | 84 | 98 |
| avg constraint (stdv) | 23.93 (8.43) | 9.91 (10.78) | 0.81 (2.86) |
| min constraint | 0 | 0 | **0** |
| max constraint | 30 | 29 | **16** |

**Table S2: E1.2: Modifying *RT:* number of simultaneously running robots, GA.**

| GA | 10 robots at once | 15 robots at once | 20 robots at once |
|---|---|---|---|
| avg objective (stdv) | 93.6 (5.55) | 91.09 (5.43) | **92.24** (4.32) |
| min objective | 76 | 80 | 80 |
| max objective | 107 | 108 | **103** |
| avg constraint (stdv) | 44.7 (3.89) | 26.69 (5.56) | **0.8** (2.97) |
| min constraint | 35 | 10 | 0 |
| max constraint | 32 | 41 | 17 |

**Table S3: E1.3: Modifying *Rb:* total number of robots, COIL.**

| COIL | 20 total robots | 25 total robots | 30 total robots |
|---|---|---|---|
| avg objective (stdv) | 50.67 (3.63) | 57.96 (7.59) | 66.43 (6.80) |
| min objective | 44 | 47 | 51 |
| max objective | 62 | 72 | 77 |
| avg constraint (stdv) | **4.4** (8.4) | 15.39 (13.13) | 23.93 (8.43) |
| min constraint | **0** | 0 | 0 |
| max constraint | **27** | 32 | 30 |

**Table S4: E1.3: Modifying *Rb:* total number of robots, GA.**

| GA | 20 total robots | 25 total robots | 30 total robots |
|---|---|---|---|
| avg objective (stdv) | 63.47 (5.33) | 81.01 (6.49) | **93.6** (5.55) |
| min objective | 51 | 64 | **76** |
| max objective | 75 | 102 | **107** |
| avg constraint (stdv) | 23.71 (7.97) | 38.35 (4.71) | 44.7 (3.89) |
| min constraint | **0** | 26 | 35 |
| max constraint | 41 | 48 | 32 |

**Table S5: E2.1: Modifying dataset size *DS***

| | 2500 data points | 5000 data points | 7500 data points | 10000 data points |
|---|---|---|---|---|
| avg objective (stdv) | 66.75 (5.87) | **68.2** (8.15) | 65.45 (8.09) | 66.43 (6.80) |
| min objective | **51** | 49 | 48 | **51** |
| max objective | 81 | **82** | 80 | 77 |
| avg constraint (stdv) | 26.52 (4.22) | 25.66 (8.02) | **23.28** (8.18) | 23.93 (8.43) |
| min constraint | 0 | 0 | 0 | **0** |
| max constraint | 31 | 33 | 31 | **30** |

**Table S6: E2.2: Modifying number of latent variables *maxlv***

| | 10 latent variables | 20 latent variables | 30 latent variables | 40 latent variables | 50 latent variables | 60 latent variables |
|---|---|---|---|---|---|---|
| avg objective (stdv) | 50.1 (2.26) | 52.08 (2.40) | 53.19 (2.86) | 53.26 (3.38) | 53.47 (3.56) | **66.43** (6.80) |
| min objective | 45 | 47 | 47 | 44 | 47 | **51** |
| max objective | 56 | 59 | 61 | 63 | 70 | **77** |
| avg constraint (stdv) | 9.25 (5.64) | **1.96** (3.03) | 4.69 (6.11) | 4.41 (6.78) | 6.69 (7.83) | 23.93 (8.43) |
| min constraint | **0** | **0** | **0** | **0** | **0** | **0** |
| max constraint | 19 | **13** | 24 | 25 | 32 | 30 |